# AI-Driven Water Segmentation with deep learning models for Enhanced Flood Monitoring

Sanjida Afrin Mou [a], Tasfia Noor Chowdhury [a], Adib Ibn Mannan[a] , Sadia Nourin Mim[b], Lubana Tarannum[b], Tasrin Noman[a], , Jamal Uddin Ahamed[*]

[a] *Department of Mechatronics & Industrial Engineering, Chittagong University of Engineering & Technology (CUET), Chattogram 4349, Bangladesh*
[b] *Department of Mechanical Engineering, Chittagong University of Engineering & Technology (CUET), Chattogram 4349, Bangladesh*

## A B S T R A C T

Flooding is a major natural hazard causing significant fatalities and economic losses annually, with increasing frequency due to climate change. Rapid and accurate flood detection and monitoring are crucial for mitigating these impacts. This study compares the performance of three deep learning models—U-Net, ResNet, and DeepLab v3—for pixel-wise water segmentation to aid in flood detection, utilizing images from drones, in-field observations, and social media. This study involves creating a new dataset that augments well-known benchmark datasets with flood-specific images, enhancing the robustness of the models. The U-Net, ResNet, and DeepLab v3 architectures are tested to determine their effectiveness in various environmental conditions and geographical locations and the strengths and limitations of each model are also discussed here, providing insights into their applicability in different scenarios by predicting image segmentation masks. This fully automated approach allows these models to isolate flooded areas in images, significantly reducing processing time compared to traditional semi-automated methods. The outcome of this study is to predict segmented mask for each image effected by flood disaster and validation accuracy of these models are DeepLab-0.9057, ResNet-0.8870, U-Net-0.8712. This methodology facilitates timely and continuous flood monitoring, providing vital data for emergency response teams to reduce loss of life and economic damages. It offers a significant reduction in the time required to generate flood maps, cutting down the manual processing time. Additionally, we present avenues for future research, including the integration of multi-modal data sources and the development of robust deep learning architectures tailored specifically for flood detection tasks. Overall, our work contributes to the advancement of flood management strategies through innovative use of deep learning technologies. All training models were uploaded to Github (**https://github.com/SanjidaAfrin25/flood-detection-using-deepLab-unet-resnet**)

## 1. Introduction

Floods are among the most disastrous natural catastrophes, inflicting major damage to infrastructure, loss of life, and economic turmoil. Early identification and monitoring play a crucial role in effective catastrophe management and mitigation. Conventional flood detection systems usually rely on manual observation, which is time-consuming and restricted. Recent advancements in deep learning algorithms open up new possibilities for automated flood detection using drone or satellite imagery.

This paper discusses the potential of deep learning methods in water segmentation for enhanced flood monitoring. Using these techniques, a possible comprehensive system can be developed that will rightfully detect the flooded zones in photos to assess actionable information for authorities and stakeholders. The increasing deployment of low-cost optical satellites, such as CubeSats, has further enabled the application of machine learning for water identification in optical and multispectral imaging (Mateo-Garcia et al.,2019; Liu Yang et al., 2015; Yang Chen et al., 2018). Yet, despite such developments, much flood analysis remains manual or semi-automated and is provided by organizations like the United Nations Institute for Training and Research - Operational Satellite Applications Program through the provision of a 'Rapid Mapping' service (Edoardo Nemni et al., 2020). Flood detection with water segmentation enhances monitoring and prediction capabilities that involve quick action, improve disaster preparedness, and support decision-making in risk management. This will not only improve public safety through early warnings but will also help in designing resilient infrastructure, land use planning, and reducing costs due to floods. The proposed method enhances the study of image processing and computer vision for community awareness and well-informed decisions.

Deep learning uses multi-layered neural networks to find patterns in data. These algorithms mimic the thought process of the human brain, making them particularly efficient in the analysis of flood imagery (Janis BARZDINS et al. 2024)**.** Integration of water segmentation with deep learning is integrally important in enhancing disaster preparedness, minimizing economic losses, and saving lives by availing accurate and efficient flood detection means. Some scholars have found a hybrid approach with robust similarity scores in the flood monitoring of Malaysia, which improves the regional flood monitoring systems (Muhadi et al., 2020) others used anisotropic diffusion segmentation and SVM to process the satellite images for disaster management, with the help of morphological operation to enhance the performance of flood monitoring (FU et al., 2010). A CNN-based method for rapid flood mapping using Sentinel-1 SAR imagery was proposed, reducing map development time by 80% and enabling accurate monitoring across diverse conditions (Nemni et al., 2020). Convolutional and recurrent neural networks were utilized to predict flash flood probability in Golestan Province, Iran, with CNN models achieving higher accuracy through geospatial databases and SWARA techniques (Panahi et al., 2021). Some efforts focused on using LBP, HOG, and pre-trained VGG-16 for floodwater detection on roadways, with VGG-16 and logistic regression showing superior performance. FCN outperformed superpixel-based methods for segmentation, enhanced further by CRF (Sarp et al., 2022). A modified U-NET model, U-FLOOD, was developed to predict 2D water depth maps in urban floods using hyetographs and topographical data, delivering fast and accurate predictions (Löwe et al., 2021). Another approach involved applying CNNs like YOLOv3 and Fast R-CNN for flood label detection with connected vision systems, integrating edge detection and aspect ratio analysis for real-time monitoring (Pally & Samadi, 2022). A fully automated end-to-end system for predicting flood stage data employed U-Net CNNs for segmentation and LSTMs for time-series prediction, achieving high accuracy in real-time forecasts (Windheuser et al., 2023). Additionally, a NN-SGW hybrid model was introduced for flood inundation mapping in data-scarce regions, identifying key environmental variables and achieving enhanced performance in urban flood prediction and susceptibility assessment (Darabi et al., 2021).

Traditional methods frequently lack the accuracy needed for effective flood detection and monitoring, resulting in inadequate mitigation strategies. Many existing solutions also struggle with scalability, making them unsuitable for diverse regions or environmental conditions. Additionally, current flood monitoring systems rely heavily on semi-automated processes, which demand considerable manual effort, reducing efficiency and increasing response times.

The objective of the study is to enhance water segmentation accuracy using advanced deep learning models, addressing challenges like scalability, manual intervention, and limited datasets by integrating diverse data sources, including satellite, drone, and social media imagery.

## 2. Materials

### 2.1. Datasets

In this study, the focus is on collecting a comprehensive dataset specifically designed for flood detection and monitoring using deep learning techniques. The dataset consists of two primary components: actual flood area images and corresponding mask images. Here, a detailed overview of the data collection process is provided.

The flood area images were collected from various online sources, including public datasets such as Kaggle, social media platforms, and open-access satellite imagery repositories (showing in Figure 1. 1 and Figure 1.2 Original mask image example. These sources were chosen to ensure a diverse set of images representing different types of flood scenarios and geographical locations. The dataset comprises 290 high-resolution images of areas affected by flooding. These images were selected to include a wide range of flood characteristics, such as urban and rural settings, different water levels, and various types of flooding, including riverine floods, flash floods, and coastal flooding.

Corresponding to each actual flood area image, a mask image was generated. These mask images were either obtained from existing annotated datasets or created manually using image annotation tools. In cases where masks were created manually, experts in the field of remote sensing and image processing annotated the water bodies in the images. Each mask image is a binary representation of the actual flood area image. Pixels representing water bodies are assigned the value 1 (white) while regions that do not hold water are assigned the value 0 (black). This binary notation is essential when training segmentation models because it endows these models with the learning capability to distinguish flooded from unflooded regions.

## 3. Methodology

The research question calls for the development of a structured work strategy or plan. In this chapter, we deduce a work plan by examining the existing literature on flood detection, water segmentation, and predicting flood effects using a deep learning model. Error! Reference source not found. presents a flowchart that visually outlines the entire working procedure for the research. By following this organized approach, we can effectively carry out the investigation and achieve our research objectives.

### 3.1. Preprocessing

Data preprocessing is vital to convert raw data into a usable format for analysis using deep learning. It involves data cleaning, normalizing numerical features, encoding categorical variables, and reducing dimensionality. The dataset is split into training and testing sets, with techniques like cross-validation ensuring robustness. For image data, augmentation is used to enhance the dataset. Handling imbalanced data through resampling or synthetic data generation ensures balanced class distributions. This process improves model accuracy, reduces complexity, and enhances generalization, leading to reliable results.

#### 3.1.1. Image Resizing

The actual images are 1024x1024 which is too bigger for the process so it should be resized to a standard dimension (256x256) to ensure uniformity across the dataset. This helps in reducing computational complexity and ensuring compatibility with the neural network input requirements.

Images

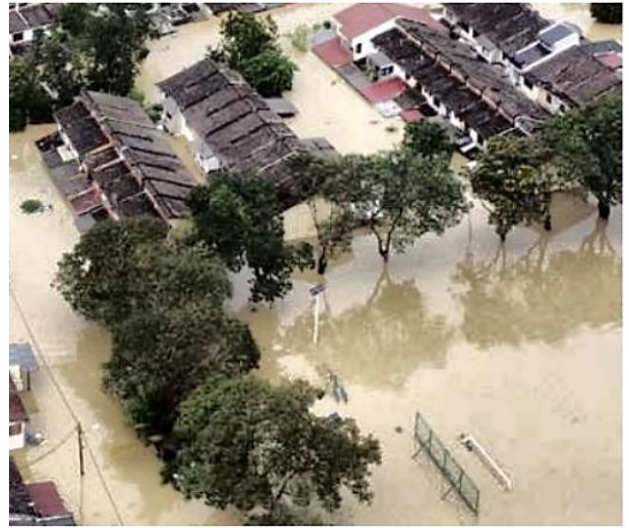

Figure 1. 1 Flood effected area

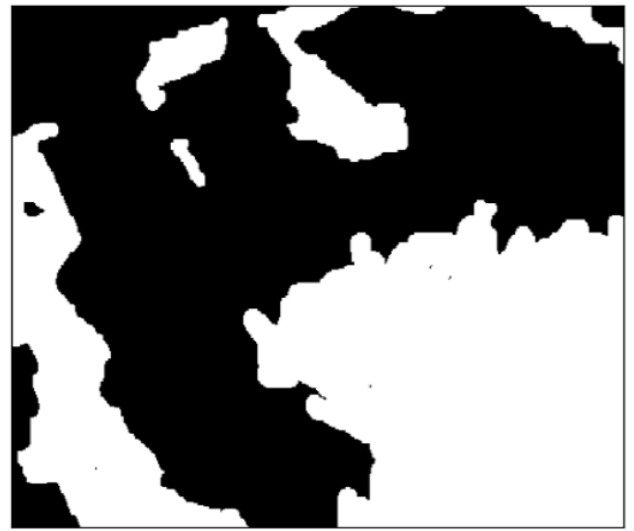

Figure 1.2 Original mask image example

#### 3.1.2. Augmentation

Various data augmentation techniques, such as rotation, flipping, blurry effect, gray scale effect, and scaling, are applied to increase the diversity of the training dataset as shown in Fig. 1. This helps in making the model more robust and generalizable and make relatable with real-world images.

#### 3.1.3. Normalization

The values are normalized on a specific range (normally 0-1) in order to facilitate a better convergence of a neural network model during the training phase.

#### 3.1.4. Convert to channel 1 image

To carry out a binary segmentation task, mask images are reformatted into single-channel. This stage makes sure that the mask images are ready for use with deep learning models, thus cutting down on the time needed for processing.

### 3.2. Combined datasets

After preprocessing, the actual and mask images are combined into a unified dataset. This combined dataset is necessary for training and testing the deep learning models, which will provide paired inputs-actual images and outputs-mask images-for supervised learning.

### 1.1. Datasets splitting

The dataset as a whole is divided into different subsets so as to test the model on new data and avoid overfitting. We utilized various data augmentation techniques to enhance the variability of the dataset. These augmentations improve the diversity of training data, enabling our deep learning model to perform more effectively and adapt to a wider range of scenarios. General ratios of cuts like 80,20 are employed. We used here Train Dataset: 80%, Validation Dataset: 20%. The set of data which has been trained is utilized for training purposes of the models while the set of data that is tested is utilized to measure their generalization ability.

### 1.2. Model selection and training

For this particular study, three deep learning models have been picked since they have been validated to perform tremendous jobs on the said area of image segmentation. The architecture for DeepLabv3, U-Net, and ResNet-50 is done using TensorFlow and Keras. These models are created with suitable layers, activation functions, and initializers.

#### 1.2.1. DeepLab v3

Deep Lab (Fig. 3) is a state-of-the-art model for semantic image segmentation. It uses an atrous convolution to capture multi-scale context by probed image at multiple sampling rates. This ability to capture context at different scales makes DeepLab highly effective for segmentation tasks.

#### 1.2.2. U-Net

U-Net (Fig. 3) is a Convolutional Network architecture for biomedical image segmentation. It contains a unique U-shaped architecture, consisting of a contracting path to capture context and a symmetric expanding path for precise localization, enabling it to segment images very effectively by capturing both low-level and high-level features.

#### 1.2.3. ResNet-50

ResNet-50 (Fig. 4) is a deep residual network with 50 layers and skip connections to prevent vanishing gradients, enabling efficient training of deep networks. Its residual blocks enhance pattern recognition, improving accuracy and efficiency. ResNet-50's architecture has inspired many models and remains foundational in deep learning advancements.

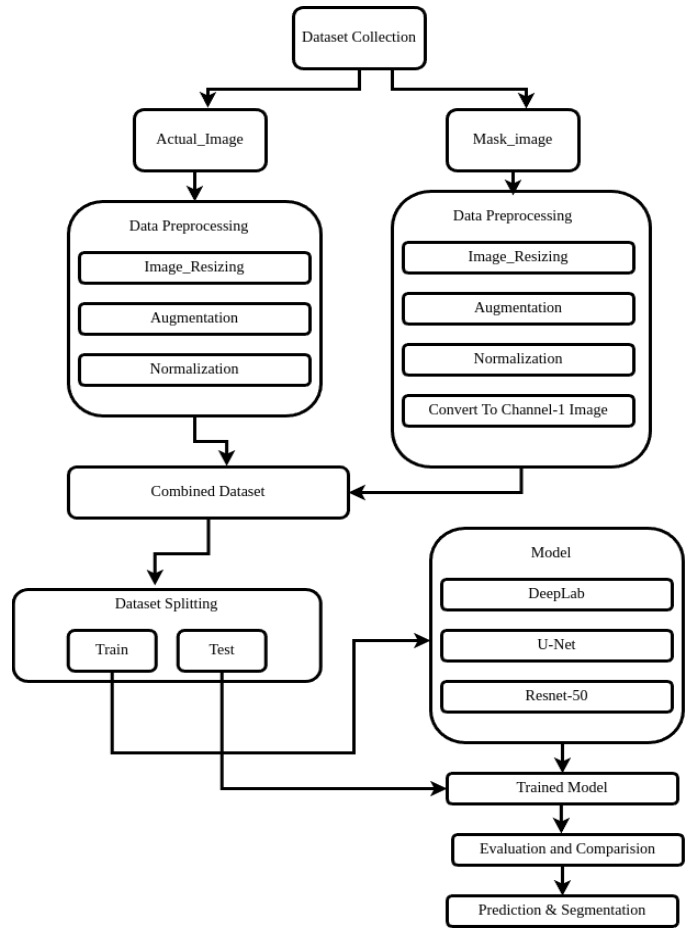

Fig. 2 Methodology flowchart of the research

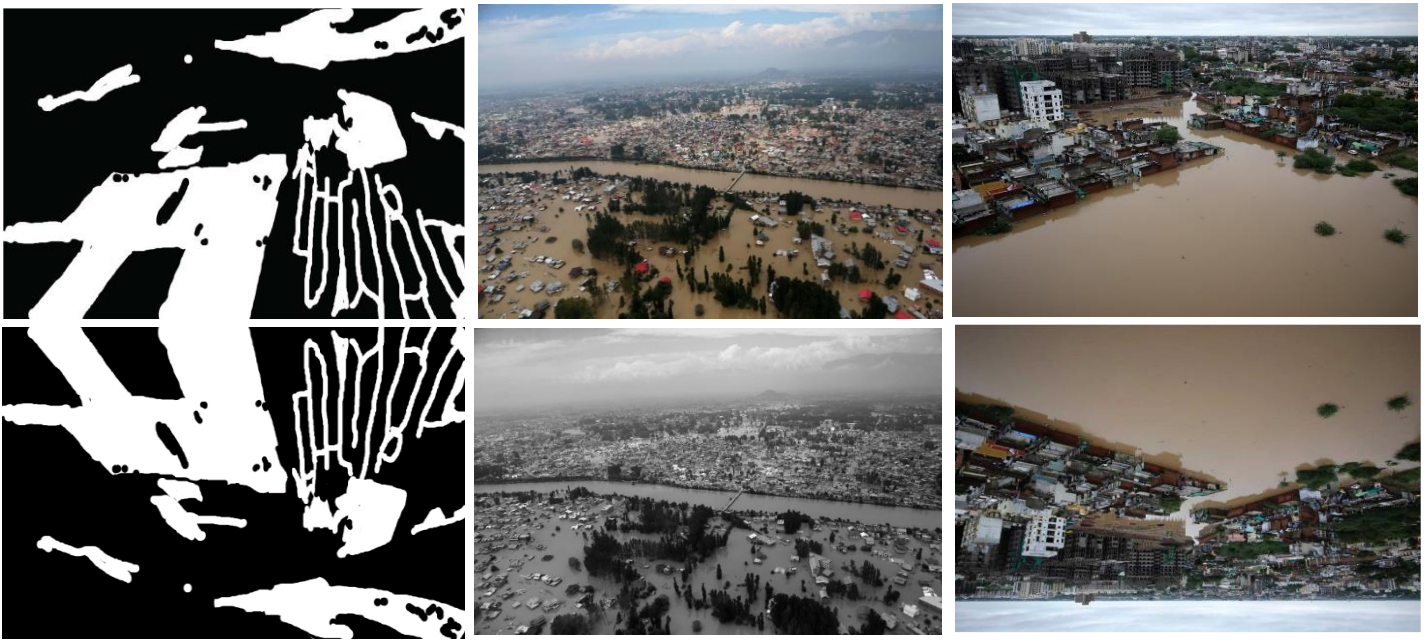

Fig. 1 Augmentation examples.

### *1.3. Evaluation Metrics and Configuration*

The performance of the deep learning models was assessed using various metrics, including accuracy, precision, recall, and F1-score, to determine their effectiveness in flood segmentation tasks. The evaluation of the models was performed on a test dataset that included diverse flood scenarios.

The metrics include validation loss, validation accuracy, precision, recall, and F1 score.

#### *1)Validation Loss*

This metric indicates how well the model performs on the validation dataset. Lower values are better and suggest that the model is not overfitting and is generalizing well.

#### *2) Validation Accuracy*

This measures the proportion of correctly predicted instances among the total instances in the validation dataset. Higher values indicate better performance.

$$Accuracy = \frac{\text{True Positives} + \text{True Negative}}{\text{True Positives} + \text{True Negative} + \text{False Positives} + \text{False Negatives}}$$

#### *3) Precision*

Precision is the ratio of true positive predictions to the summation of true positive and false positive predictions. It measures the accuracy of positive predictions.

$$Precision = \frac{\text{True Positives}}{\text{True Positives} + \text{False Positives}}$$

Early stopping helps to find the optimal number of epochs for on the specific training images and struggles to identify floods in new, unseen data. The Adam algorithm (Adaptive Moment Estimation) works alongside epochs and early stopping during the training process of flood detection training by preventing overfitting and ensuring the models generalize well to new flood images. If the validation performance does not improve for a specified number of epochs (defined by the patience parameter), early stopping is triggered.

#### *4) Recall*

Recall, also known as sensitivity or true positive rate, The ratio of true positive predictions to the total of true positive and false negative predictions. It measures the ability of the model to find all relevant instances.

$$Recall = \frac{\text{True Positives}}{\text{True Positives} + \text{False Negatives}}$$

#### *5) F1 Score*

The F1 score is calculated by taking the harmonic mean of recall and precision. Both false positives and false negatives are considered in this fair metric.

$$\text{F1 Score} = 2 \times \frac{Precision \times Recall}{\text{Precision} + \text{Recall}}$$

### *1.4. Optimization*

In training deep learning models for flood detection to learn how to identify flood pixels from various image sources (drones, field observations, social media), three key techniques (epoch, adam optimizer, early stopping) work together to achieve the best possible results. The first is epochs, which represent how many times the entire training dataset (images containing floods) is shown to the model. More epochs allow the model to learn more intricate patterns in the data, like the subtle differences between water and land. However, too many epochs can lead to overfitting, where the model becomes overly focused.

## 2. Result

**Fig. 7** demonstrates the water segmentation results from three deep learning models: DeepLabv3, U-Net, and ResNet. While all models produced satisfactory segmentation masks, DeepLabv3 performed better in distinguishing between water and non-water bodies. In terms of accuracy, DeepLabv3 achieved the highest accuracy at 90.57%, followed by ResNet at 88.7%, and U-Net at 87.12%. The accuracy was calculated using the formula:

$$Accuracy = \frac{\text{TP} + \text{TN}}{\text{TP} + \text{TN} + \text{FP} + \text{FN}}$$

Here, True Positive (TP) represents pixels correctly identified as "water," True Negative (TN) represents pixels correctly identified as "others," False Positive (FP) refers to "water" pixels mislabeled as "others," and False Negative (FN) refers to "others" pixels mislabeled as "water." This analysis highlights DeepLabv3's superior performance for water segmentation tasks.

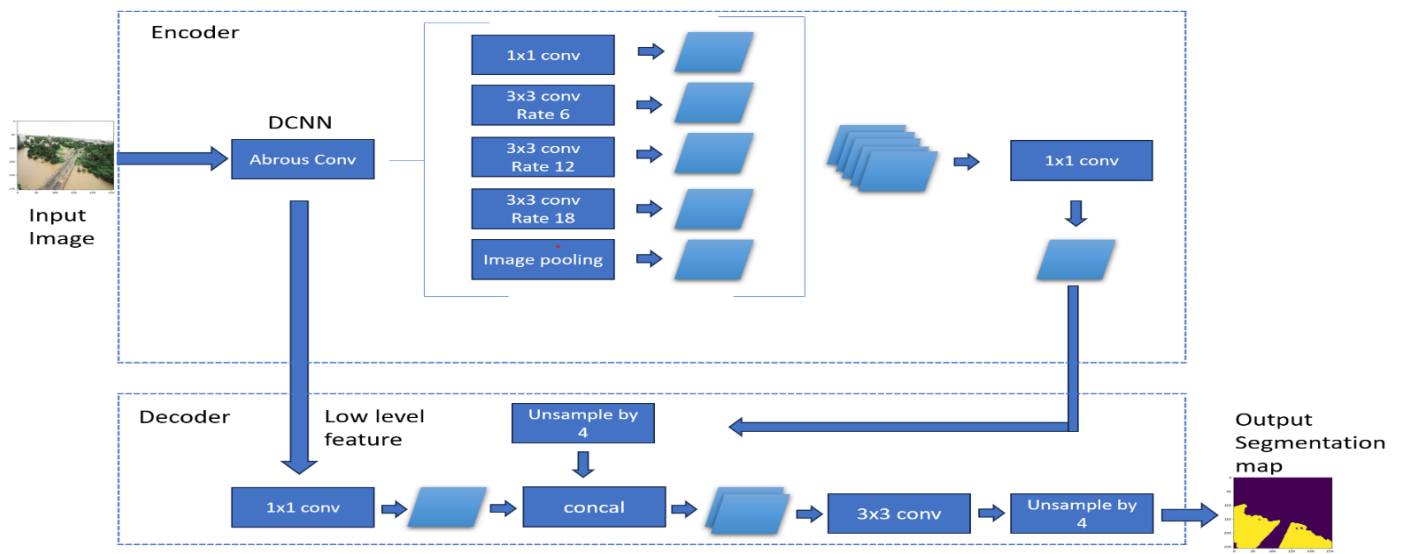

Fig. 2 DeepLab v3 architecture

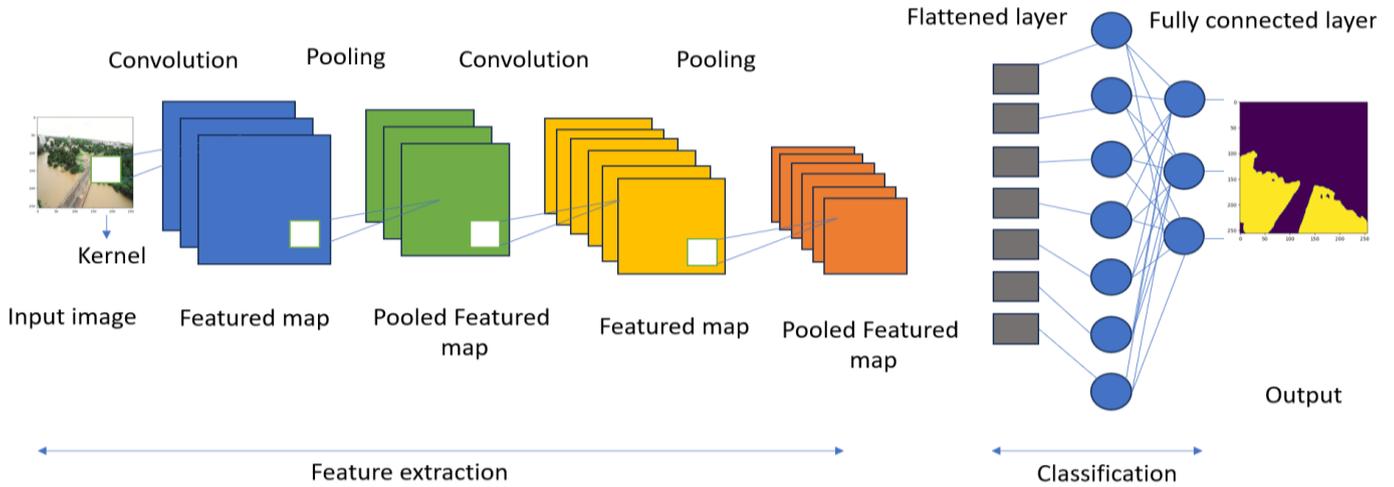

Fig. 3 U-Net architecture.

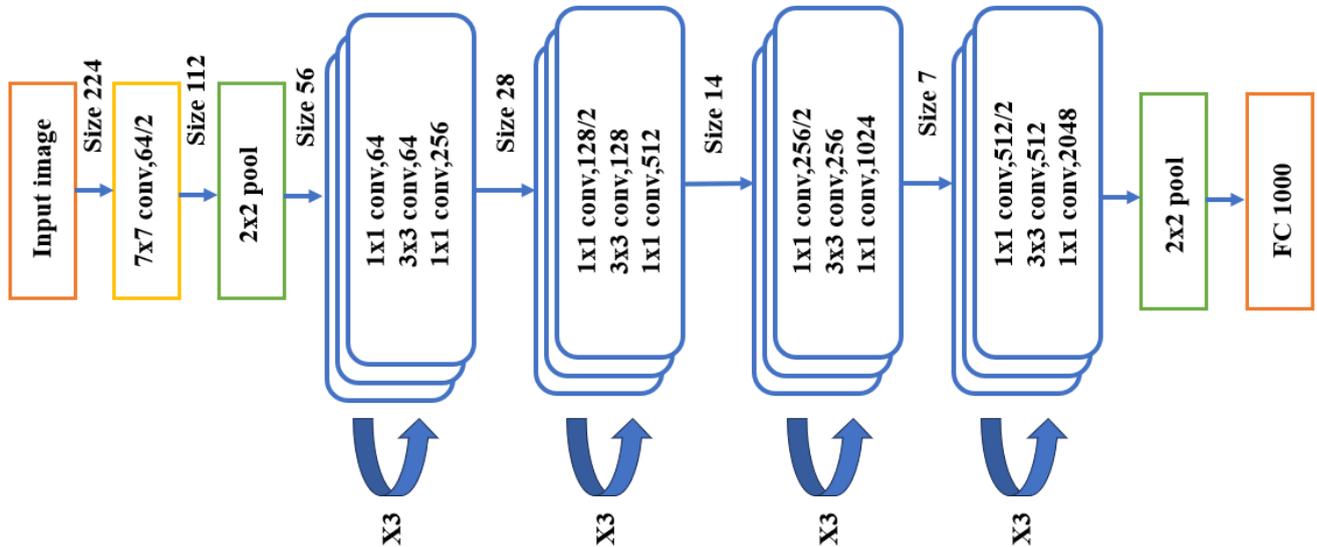

Fig. 4 Res-Net architecture

---

**Algorithm 1: Flood Area Segmentation using DeepLabV3, UNet, and ResNet50**

---

1. **Import** libraries and set dataset paths for images, masks, and class folders.
2. **Define** root folder and dataset paths.
3. **Load** the dataset.
4. **Perform** preprocessing the dataset by resizing, augmenting and normalizing images and masks, and split it into training, validation, and test sets.
5. **Define** PyTorch DataLoaders for efficient data handling and batching.
6. **Implement** model architecture:
   - **DeepLabV3**: Initialize with ResNet backbone and ASPP.
   - **UNet**: Implement encoder-decoder with skip connections.
   - **ResNet50**: Use pre-trained ResNet50 with custom segmentation head.
7. **Move** the selected model to the device (CPU/GPU).
8. **Define** loss function (Dice Loss or Binary Cross-Entropy) and optimizer (Adam/SGD).
9. **Train** the model by looping through epochs, performing forward pass, calculating loss, backpropagating, and validating.
10. **Save** the training history.
11. **Save** the model and optimizer state after training.
12. **Evaluate** the model on test data, performing predictions and calculating metrics (IoU, Dice Score).
13. **Plot** training and validation metrics over epochs.
14. **Save** and visualize results by displaying input images, ground truth, and predicted masks.

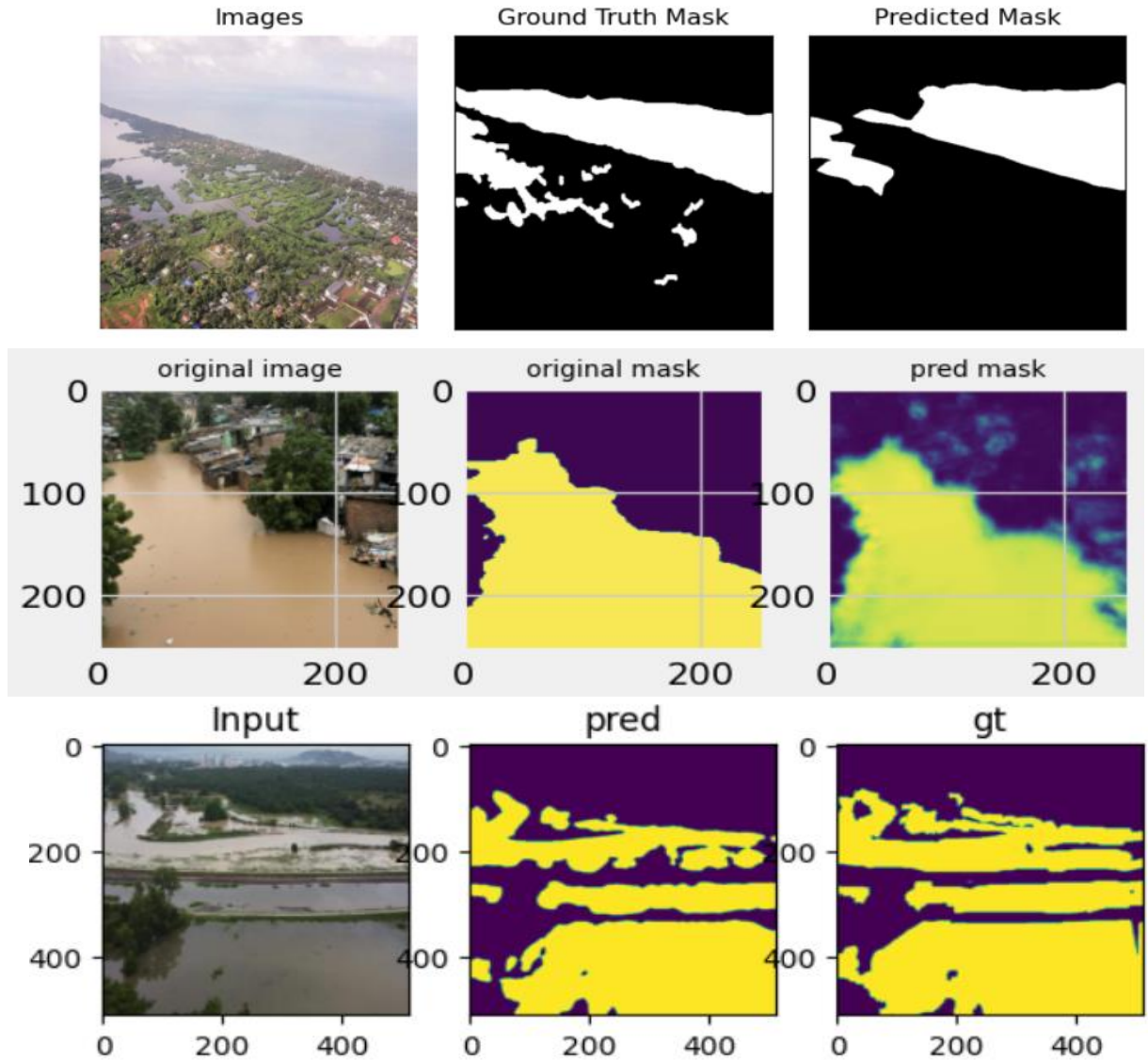

Fig. 5 Visualization of Predicted Flood Segmentation Results for DeepLabv3 (Top Row), U-Net (Middle Row), and ResNet-50 (Bottom Row)

## 3. Discussion

The models were evaluated based on their ability to accurately segment flood-affected areas. Metrics such as accuracy, precision, recall, and F1-score were used to quantify the performance.

**Table 1**

Performance Comparison

| Methods | DeepLabv3 | U-Net | ResNet |
|---|---|---|---|
| **Validation loss** | 0.0314 | 0.0240 | 0.0251 |
| **Accuracy** | 0.9057 | 0.8712 | 0.8870 |
| **Precision** | 0.8840 | 0.8200 | 0.8670 |
| **Recall** | 0.8707 | 0.8414 | 0.8460 |
| **F1 Score** | 0.8749 | 0.8361 | 0.8510 |

Error! Reference source not found. presents a variety of performance measures for deep learning models. U-Net, ResNet, and DeepLab models show sufficient accuracy and reliability for flood detection and monitoring, indicating wide applicability of other approach. The capabilities of segmented reconstruction of depths under different conditions also outperformed the quality of traditional methods. The results suggest that deep learning has a great potential to improve disaster management by facilitating quicker action and intervention during response to floods. Although DeepLabv3 shows the highest validation loss, it still shows excellent performance in terms of general accuracy and precision. On the other hand, U-Net has the lowest validation loss, indicating good generalization of this model on the validation set. However, it has less accuracy and a lower F1 score compared to the other two models, which implies that it would be less reliable in predicting positive cases as compared to DeepLabv3 and Res-Net. ResNet is balanced with a low validation loss, high validation accuracy, and acceptable precision and recall values. It obtains the second-best F1 score, placing it as a strong model for flood segmentation tasks. Figure 8: "Training and Validation Loss" on the left and "Training and Validation Accuracy" on the right, both plotted against epochs for the three models: DeepLabv3, U-Net, and ResNet-50. The following charts elucidate the learning behavior of each model during training. DeepLabv3 shows stable convergence with the lowest validation loss and highest validation accuracy; hence, it generally performs well and indicates effective generalization. In contrast, U-Net reveals the lowest training loss, although it has a bit higher validation loss, which might cause overfitting. ResNet-50 presents balanced performances in terms of accuracy and loss, remaining stable and becoming one of the robust models for the flood segmentation task.

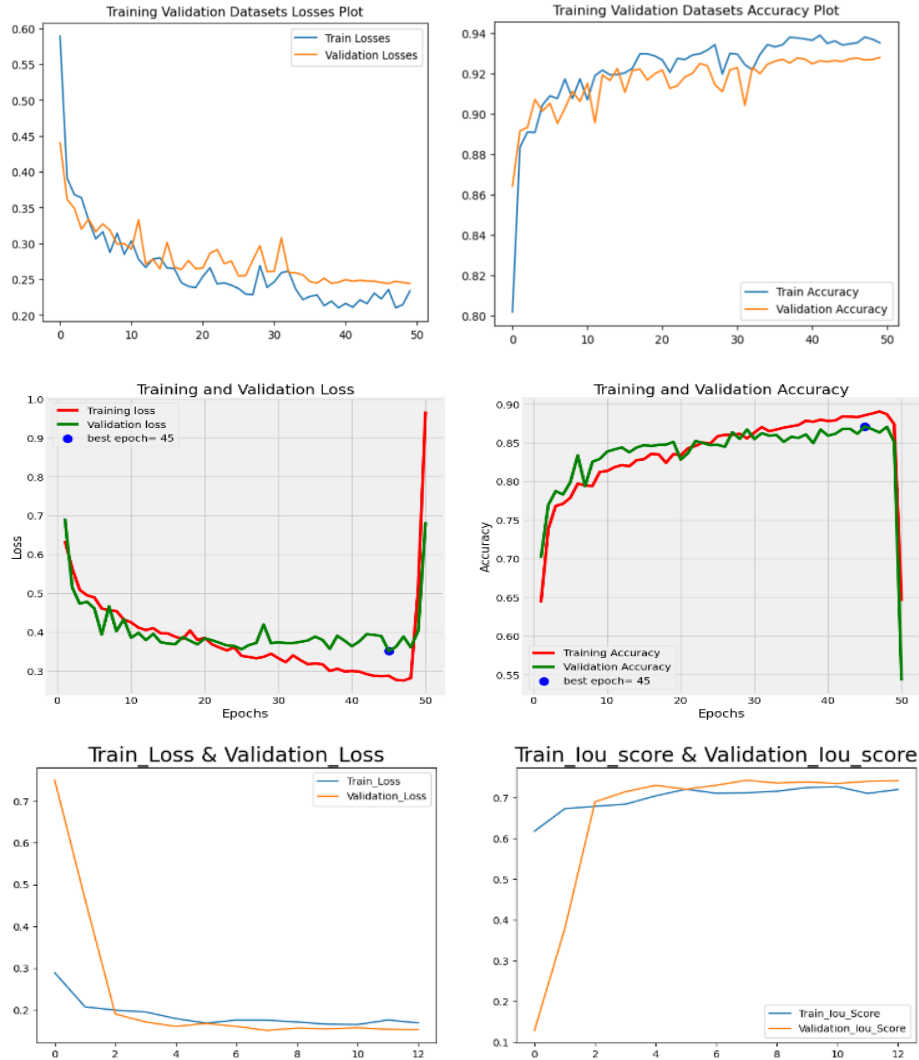

Fig. 6 "Training and Validation Loss" on the left and "Training and Validation Accuracy" on the right, both plotted against epochs for 3 models (DeepLabv3, U-Net and Res-Net50)

## 4. Conclusion

In conclusion, this work highlights the potential of deep learning models—U-Net, ResNet, and DeepLabv3—for effective and efficient flood detection and monitoring. This automation method substantially shortens the time required for flood mapping and improves precision in identifying flooded areas. The enhanced dataset, complemented by detailed model evaluation, provides valuable resources for future research efforts. All deep learning segmentation models show average results for segmentation however, DeepLabv3 shows that it has the highest validation accuracy and a better F1 score, which makes it the best model for flood segmentation tasks among the three.

U-Net has the lowest validation loss, indicating the best generalization to the validation dataset, however, it performs lower in accuracy and F1 score. Overall ResNet has a good balance in terms of parameters such as the validation loss, accuracy, precision, recall, and F1 score. This comprehensive review provided some degree of insight into the advantages and drawbacks of each model along with recommendations for future improvements and implementations. The study shows that flood detection can be achieved reliably without a costly setup, making it achievable for wider implementation in resource-constrained contexts. By automating the flood detection process, the technology decreases the possibility for human error in manual flood mapping, resulting to more accurate and dependable findings.

Future advancements in flood detection and monitoring systems are likely to involve significant innovations in the integration of multi-modal data sources, such as the combination of optical, radar, and satellite-based thermal imaging. This fusion of diverse datasets would enhance the robustness and accuracy of models by leveraging the strengths of each modality—for instance, radar's ability to penetrate cloud cover and optical imaging's high spatial resolution. Simultaneously, progress in AI and machine learning will allow the creation of AI algorithms that can process huge quantities of data more efficiently. Such systems could detect patterns and anomalies within real time data to allow for faster detection and prediction of flooding events. Better computational infrastructure, including but not limited to edge computing and cloud-based solutions, may significantly reduce data processing time, enabling real-time monitoring and the rapid distribution of early warning systems to impacted areas.

Moreover, a network of Internet of Things (IoT) devices, such as ground-based sensors for measuring water levels and rainfall intensity if sare harnessed would complement satellite data, creating a comprehensive and interconnected monitoring network. Future systems may also utilize crowd-sourced data — such as social media posts and photos — to enhance situational awareness and validation on the ground. In addition to detecting imminent threats, emerging communication technologies, such as 5G and low Earth orbit (LEO) satellite networks will allow alerts to be disseminated quickly to the affected communities, emergency response teams, and government agencies. These systems could also include predictive analytics to model the possible ripple effect of flooding, supporting better resource management and evacuation planning.

By combining these technological innovations, the next generation of flood detection and monitoring systems will be more proactive, accurate, and responsive, ultimately reducing the devastating impacts of floods on lives and infrastructure.

### Credit authorship contribution statement

**S.A Mou, T.N Chowdhury:** Conceptualization, Methodology, Formal analysis, Validation**, Adib I.M, S.N Mim:** Investigation, Resource refinement, data curation. **L. Tarannum, T. Noman:** Software, data curation. **J. U. Ahamed:** Supervision, Project Administration. All authors contributed to the revision of the manuscript and have read and approved its final version.

### Declaration of Competing Interest

The authors declare that they have no known competing financial interests or personal relationships that could have appeared to influence the work reported in this paper.

### References

[1] Mateo-Garcia, G.; Oprea, S.; Smith, L.J.; Veitch-Michaelis, J.; Schumann, G.; Gal, Y.; Baydin, A.G.; Backes, D. Artificial Intelligence for Humanitarian Assistance and Disaster Response Workshop, 33rd Conference on Neural Information Processing Systems (NeurIPS), Vancouver, Canada. arXiv 2019, arXiv:1910.03019

[2] Yang, L.; Tian, S.; Yu, L.; Ye, F.; Qian, J.; Qian, Y. Deep Learning for Extracting Water Body from Landsat Imagery. Int. J. Innov. Comput. Inf. Control 2015, 11, 1913–1929.

[3] Chen, Y.; Fan, R.; Yang, X.; Wang, J.; Latif, A. Extraction of Urban Water Bodies from High-Resolution Remote-Sensing Imagery Using Deep Learning. Water 2018, 10, 585. [CrossRef]

[4] UNITAR's Operational Satellite Applications Programme—UNOSAT. Rapid Mapping Service. Available online: https://www.unitar.org/maps/unosat-rapid-mapping-service (accessed on 4 August 2020).

[5] UNITAR's Operational Satellite Applications Programme—UNOSAT. Rapid Mapping Service. Available online: https://www.unitar.org/maps/unosat-rapid-mapping-service (accessed on 4 August 2020).

[6] Introduction to Deep Learning Studio—Deep Learning Studio | Documentation. (n.d.). https://doc.arcgis.com/en/deep-learning-studio/latest/get-started/about-deep-learning-studio.htm

[7] N. Muhadi, A. Abdullah, S. Bejo, M. Mahadi, and A. Mijic, "Image Segmentation Methods for Flood Monitoring System," *Water*, vol. 12, no. 6, p. 1825, Jun. 2020, doi: 10.3390/w12061825.

[8] J. FU, Q. SU, S. PAN, and J. LU, "Support Vector Machine Based Groundwater Level Monitoring Model by Using Remote Sensing Images," *Geo-information Science*, vol. 12, no. 4, pp. 466–472, Aug. 2010, doi: 10.3724/sp.j.1047.2010.00466.

[9] E. Nemni, J. Bullock, S. Belabbes, and L. Bromley, "Fully Convolutional Neural Network for Rapid Flood Segmentation in Synthetic Aperture Radar Imagery," *Remote Sensing*, vol. 12, no. 16, p. 2532, Aug. 2020, doi: 10.3390/rs12162532.

[10] M. Panahi et al., "Deep learning neural networks for spatially explicit prediction of flash flood probability," *Geoscience Frontiers*, vol. 12, no. 3, p. 101076, May 2021, doi: 10.1016/j.gsf.2020.09.007.

[11] Sarp, S., Kuzlu, M., Zhao, Y., Cetin, M., & Guler, O. (2022). A comparison of deep learning algorithms on image data for detecting floodwater on roadways. *Computer Science and Information Systems*, *19*(1), 397–414. https://doi.org/10.2298/csis210313058s

[12] Löwe, R., Böhm, J., Jensen, D. G., Leandro, J., & Rasmussen, S. H. (2021, December). U-FLOOD – Topographic deep learning for predicting urban pluvial flood water depth. *Journal of Hydrology*, *603*, 126898. https://doi.org/10.1016/j.jhydrol.2021.126898

[13] Pally, R., & Samadi, S. (2022, February). Application of image processing and convolutional neural networks for flood image classification and semantic segmentation. *Environmental Modelling & Software*, *148*, 105285. https://doi.org/10.1016/j.envsoft.2021.105285

[14] Windheuser, L., Karanjit, R., Pally, R., Samadi, S., & Hubig, N. C. (2023, January). An End-To-End Flood Stage Prediction System Using Deep Neural Networks. *Earth and Space Science*, *10*(1). https://doi.org/10.1029/2022ea002385

[15] Darabi, H., Torabi Haghighi, A., Rahmati, O., Jalali Shahrood, A., Rouzbeh, S., Pradhan, B., & Tien Bui, D. (2021, December). A hybridized model based on neural network and swarm intelligence-grey wolf algorithm for spatial prediction of urban flood-inundation. *Journal of Hydrology*, *603*, 126854. https://doi.org/10.1016/j.jhydrol.2021.126854

[16] *Kaggle: your machine learning and data science community*. (n.d.). https://www.kaggle.com/